# Evaluating Indirect Strategies for Chinese–Spanish Statistical Machine Translation

**Marta R. Costa-jussà**                                      VISMRC@I2R.A-STAR.EDU.SG
*Institute for Infocomm Research,*
*Singapore 138632*

**Carlos A. Henríquez Q.**                                   CARLOS.HENRIQUEZ@UPC.EDU
*Universitat Politècnica de Catalunya,*
*08034 Barcelona*

**Rafael E. Banchs**                                         REMBANCHS@I2R.A-STAR.EDU.SG
*Institute for Infocomm Research,*
*Singapore 138632*

## Abstract

Although, Chinese and Spanish are two of the most spoken languages in the world, not much research has been done in machine translation for this language pair. This paper focuses on investigating the state-of-the-art of Chinese-to-Spanish statistical machine translation (SMT), which nowadays is one of the most popular approaches to machine translation. For this purpose, we report details of the available parallel corpus which are Basic Traveller Expressions Corpus (BTEC), *Holy Bible* and United Nations (UN). Additionally, we conduct experimental work with the largest of these three corpora to explore alternative SMT strategies by means of using a pivot language. Three alternatives are considered for pivoting: cascading, pseudo-corpus and triangulation. As pivot language, we use either English, Arabic or French. Results show that, for a phrase-based SMT system, English is the best pivot language between Chinese and Spanish. We propose a system output combination using the pivot strategies which is capable of outperforming the direct translation strategy. The main objective of this work is motivating and involving the research community to work in this important pair of languages given their demographic impact.

## 1. Introduction

Chinese and Spanish are very distant languages in many aspects. However, they come close together in the ranking of most spoken languages in the world (Ethnologue, 2012). In the Web 2.0 era, in which most of the content is produced by the users, the number of native speakers is an excellent indicator of the actual relevance of machine translation between two languages. Of course, other factors such as literacy, amount of text published and strength of commercial relationships are also to be taken into account, but these factors will actually support further our idea of the strategic importance of developing machine translation technologies between Chinese and Spanish. The huge increase in volume of online contents in Chinese during the last years, as well as the steady increase of commercial relationships between Spanish speaking Latin American countries and China are just two basic examples supporting this fact. Needless to say, these languages involve many economical interests (Zapatero, 2010). Nevertheless, these two languages seem to become far apart again when looking for bilingual resources.

We have been recently interested in gathering and collecting Chinese–Spanish bilingual resources for research and machine translation application purposes. The amount of bilingual resources that are currently available for this specific language pair is surprisingly low. Similarly, the related amount of work we have found, within the computational linguistic community, can be reduced to a very small set of references (Banchs, Crego, Lambert, & Mariño, 2006; Banchs & Li, 2008; Bertoldi, Cattoni, Federico, & Barbaiani, 2008; Wang, Wu, Hu, Liu, Li, Ren, & Niu, 2008). Apart from the BTEC[1] corpus available through International Workshop on Spoken Language Translation (IWSLT) competition (Bertoldi et al., 2008) and *Holy Bible* datasets (Banchs & Li, 2008), we were not aware of any other Chinese–Spanish parallel corpus suitable for training phrase-based (Koehn, Och, & Marcu, 2003)[2] statistical machine translation systems between these two languages, until a six-language parallel corpus (including both Chinese and Spanish) from United Nations was released for research purposes (Rafalovitch & Dale, 2009).

Using the recently released United Nations parallel corpus as a starting point, this work focuses on the problem of developing Chinese-to-Spanish phrase-based machine translation technologies with a limited set of bilingual resources. We explore and evaluate different alternatives for the problem in hand by means of pivot-language strategies through other languages available in the United Nations parallel corpus, such as Arabic, English and French [3]. Existing strategies such as system cascading, pseudo-corpus generation and triangulation are implemented and compared against a baseline system built with a direct translation approach. As follows, we briefly describe these pivot approaches:

- The cascaded approach generates Chinese-to-Spanish translations by concatenating a system that translates Chinese into a pivot language with a system that translates from the pivot language into Spanish.

- The pseudo-corpus approach builds a synthetic Chinese–Spanish corpus either by translating into Spanish the pivot side of a Chinese–pivot corpus or by translating into Chinese the pivot side of a Pivot–Spanish corpus.

- The triangulation approach implements a Chinese-to-Spanish translation system by combining the translation table probabilities of a Chinese–pivot system and a Pivot–Spanish system.

Additionally, we implement and evaluate a system combination of the three pivot strategies based on the minimum Bayes risk (MBR) (Kumar & Byrne, 2004) technique. Such a combination strategy is capable of outperforming the direct system.

Besides experimenting with different pivot languages to compare the mentioned approaches, we also wanted to determine which pivot alone gives the best results and why.

---

1. Basic Traveller Expressions Corpus.

2. Note that *phrase-based* is commonly used to refer to statistical machine translation systems, in which the term *phrase* refers to segments of one or more than one word and it does not have the usual meaning of *multi-word syntactical consitutent,* as it has in linguistics.

3. Although Russian is available in the UN corpus, we discard to use it because we do not have the proper preprocessing tools for it.

Hence, we present a short comparison of the amount of reordering and vocabulary sizes of pivot languages, following the study presented by Birch et al. (2008) where they identified these two properties as key elements for predicting machine translation quality. The results from such comparisons, together with the translation quality obtained in the different approaches, show that English was the best pivot language for Chinese-to-Spanish translation purposes in our experimental framework.

The paper is structured as follows. Section 2 motivates this work which is intended to bring some light into the investigation of Chinese-to-Spanish translation task. Section 3 presents some related work in the Chinese-to-Spanish translation task. Section 4 reports the details of the main parallel corpora available for this translation task. Next, section 5 describes the main strategies for performing Chinese-to-Spanish translation which are tested in this work: direct, cascade, pseudo-corpus and triangulation. Section 6 presents the evaluation framework which includes the corpus statistics, the system and evaluation details. Then, section 7 reports the experiments (including the system combination) and the results. Finally, section 8 concludes our work and proposes new research directions in the area.

## 2. Motivation

Although some current web translation systems allow for performing translations between Chinese and Spanish, the quality of current Chinese-to-Spanish translations is still well below the quality achieved for other language pairs, such as English to Spanish. As far as we know, there is not much research in this translation task. The main reason may be the lack of parallel corpora. This study intends to make progress and involve other researchers in the area of Chinese–Spanish statistical machine translation by:

1. Listing the available parallel corpora for Chinese–Spanish.

2. Comparing different methodologies for performing statistical machine translation: cascaded (Wang et al., 2008), pseudo-corpus generation (Banchs et al., 2006; de Gispert & Mariño, 2006) and triangulation (Wu & Wang, 2007).

3. Evaluating which is the best language (among Arabic, English and French) for generating the cascade, pseudo-corpus or triangulation MT between Chinese–Spanish.

4. Performing an output system combination to explore new ways of improving Chinese-to-Spanish translation.

## 3. Related Work

One of the first works dealing with Chinese–Spanish statistical machine translation was presented by Banchs et al. (2006). Authors experimented with two independent corpora Chinese–English and English–Spanish to translate from Chinese to Spanish. They built their translation systems using the so-called Ngram-based approach, which differs from the phrase-based system mainly in the translation and reordering model (Mariño, Banchs, Crego, de Gispert, Lambert, Fonollosa, & Costa-jussà, 2006).

The only research event recently performed for this language pair was the 2008 Iwslt evaluation campaign (Paul, 2008). This evaluation organized two Chinese-to-Spanish tracks. One of them focused on direct translation and the other one on pivot translation through English. The best translation results accordingly to the manual evaluation were obtained by far in the pivot task.

The best systems in both tracks were developed by Wang et al. (2008). Regarding the direct system, they used a standard phrase-based Smt system. What makes it different from the other participating systems is that they provide their own Chinese segmentation and the Ldc (Linguistic Data Consortium) bilingual dictionary. Regarding the pivot task, they compared two different approaches. The first one, referred to as triangulation, consisted of training two translation models on the Chinese–English corpus and English–Spanish corpus, and then building a new translation model for Chinese–Spanish translation by combining the two previous models as proposed by Wu & Wang (2007); the second one obtained better results and it was based on a cascaded approach. The idea here is to translate from Chinese into English and then from English into Spanish, which means performing two translations.

Other participants also proposed the cascaded methodology. This approximation can be done with the $n$-best translations (Khalilov, Costa-Jussà, Henríquez, Fonollosa, Hernández, Mariño, Banchs, Chen, Zhang, Aw, & Li, 2008).

Another proposal was to generate pseudo-corpus which means to translate either the English into Chinese or into Spanish, creating a parallel Chinese–Spanish corpus. This pseudo-corpus is used to train the Chinese–Spanish translation (Bertoldi et al., 2008).

As mentioned aboved, the comparison performed by Wang et al. (2008) showed that the cascaded approach performed better than the phrase-table combination for the Chinese–Spanish pivot task.

Finally, our previous work (Costa-jussà, Henríquez, & Banchs, 2011b) compared two standard pivot approaches (pseudo-corpus and cascaded) using English and the direct system. Experiments in this work showed that the quality between the direct system and the pivot systems did not differ much. Additionally, the cascaded system presented slightly better results than the pseudo-corpus system. In our other previous work (Costa-jussà, Henríquez, & Banchs, 2011a), we compared again two pivot approaches (pseudo-corpus and cascaded) using Arabic, French and English as pivot languages and the direct system. We concluded that English was the best pivot language.

In the present work, we are extending the two previous studies by: (1) using more pivot strategies (including the triangulation strategy); (2) introducing a measure to pre-evaluate the quality of pivot approaches; (3) extending the pivot combination experiments; and (4) providing further evaluation.

Note that we are working with the United Nations (Un) corpus rather than with the Btec corpus (the one used in the Iwslt). The former is freely available and larger than the latter.

## 4. Chinese–Spanish Parallel Corpora

There are very limited resources for the language pair Chinese–Spanish in comparison to the number of native speakers in these languages. In practice, it is also common to translate Chinese into Spanish through English even when manual translations are conducted.

As parallel corpus at the sentence level, there is the Basic Travel Expressions Corpus (Btec) (Paul, Yamamoto, Sumita, & Nakamura, 2009), which is a collection of sentences that bilingual travel experts consider useful for people going to or coming from another country. This corpus contains around 160,000 parallel sentences but only around 20,000 sentences and 180,000 words are actively used for Mt purposes in the Iwslt evaluation campaign. The full corpus is not freely available, and the 20,000 version was only available for participation purposes in the 2008 Iwslt evaluation campaign.

Another parallel corpus is the *Holy Bible*, which has been proved to be a good resource for CLIR (Cross-language information retrieval) (Chew, Verzi, Bauer, & McClain, 2006). This corpus contains around 28,000 parallel sentences and around 800,000 tokens per language. The main advantages of using this corpus is that it is the world's most translated book; it covers a variety of literary styles including narrative, poetry, and correspondence; great care is taken over the translations; and, perhaps surprisingly, its vocabulary appears to have a high rate of coverage (as much as 85%) of modern-day language.

Finally, there is the United Nations multilanguage corpus (Rafalovitch & Dale, 2009), which is freely available online for research purposes. Among others, it contains parallel texts at the sentence level in the following languages: Chinese, English, Spanish, French and Arabic. It consists of 2100 United Nations General Assembly resolutions with translation in the six official languages of the United Nations, with average of around 3 million tokens per language. This is the material that we are using in this work. Table 1 shows the statistics of the three different corpora with their corresponding languages.

| Corpus | Lang. | Sent. | Words | Vocab. | Avg. sent. length |
|--------|-------|-------|-------|--------|-------------------|
| Btec | Chinese | 20 | 164 | 8 | 6 |
|  | English | 20 | 182 | 8 | 7 |
|  | Spanish | 20 | 147 | 17 | 9 |
| *Holy Bible* | Chinese | 30 | 814 | 13 | 26 |
|  | English | 30 | 908 | 12 | 29 |
|  | Spanish | 30 | 836 | 27 | 27 |
| Un | Chinese | 60 | 1,750 | 18 | 28 |
|  | English | 60 | 2,080 | 15 | 34 |
|  | Spanish | 60 | 2,380 | 20 | 39 |
|  | Arabic | 60 | 2,720 | 17 | 44 |
|  | French | 60 | 2,380 | 18 | 39 |

Table 1: Available corpora for Chinese–Spanish (all figures are given in thousands, except the average sentence length)

Additionally, we can surf the web and find several publications which are available both in Chinese and Spanish e.g. Global Asia Magazine (2012), but this additional material consists mainly of comparable corpora rather than parallel corpora. This comparable material cannot directly be used in a statistical machine translation system. However, there are many nice algorithms which can extract parallel corpora from comparable corpora (Moore, 2002; Senrich, 2010; Abdul-Rauf, Fishel, Lambert, Noubours, & Sennrich, 2012).

## 5. Direct and Pivot Statistical Machine Translation Approaches

There are several strategies that we can follow when translating a pair of languages in statistical machine translation (SMT). In this section we present the details of the ones we are using in this work.

In general, a statistical machine translation system relies on the translation of a source language sentence $s$ into a target language sentence $\hat{t}$. Among all possible target language sentences $t$ we choose the one with the highest probability, as show in equation (2):

$$\hat{t} = \arg\max_t \left[ P\left(t|s\right) \right] \tag{1}$$

$$= \arg\max_t \left[ P\left(t\right) P\left(s|t\right) \right] \tag{2}$$

This probability decomposition based on Bayes' theorem is known as the source-channel approach to statistical machine translation (Brown, Cocke, Della Pietra, Della Pietra, Jelinek, Lafferty, Mercer, & Roossin, 1990). It allows to model independently the target language model $P\left(t\right)$ and the source translation model $P\left(s|t\right)$. On the one hand, the translation model weights how likely words in the foreign language are translation of words in the source language; the language model, on the other hand, measures the fluency of hypothesis $\hat{t}$. The search process is represented as the arg max operation.

Later on, a variation was proposed by Och & Ney (2002) named log-linear model. It allows using more than two models or features and to weight them independently as can be seen in equation (3):

$$\hat{t} = \arg\max_t \left[ \sum_{m=1}^{M} \lambda_m h_m(s,t) \right] \tag{3}$$

This equation should be interpreted as a maximum-entropy framework. We see that eq. (2) is a special case of eq. (3). In fact, it is the logarithm of (2) which would be similar to (3). Then, we have to identify $h_1(s,t)$ with $\log(p(t))$ and $h_2(s,t)$ with $\log(p(s|t))$, and taking $M = 2$ (two models) and $\lambda_1 = \lambda_2 = 1$. In the general case, $\lambda$'s are obtained by maximizing an objective function on a held-out set (development set).

Among the additional features that can be used with the log-linear model we have lexical models, word bonus, and the reordering model. The lexical models are particularly useful in cases where the translation model may be sparse. For example, for phrases which may have appeared few times the translation model probability may not be well estimated. Then, the lexical models provide a probability among words (Koehn et al., 2003). The word bonus is used to compensate the language model which benefits shorter outputs. The reordering model is used to provide reordering between phrases. If not, reordering would only be treated internally in each phrase. Finally, it should be mentioned that the name *log-linear* is clearly a misnomer as many of these features are not logarithms at all.

As regards the reordering model, the standard way of implementing it is with a distance-based model that gives a linear cost depending on the reordering distance. For instance, if consecutive target words $t_1, t_2$ come from translating source words $s_1$ and $s_5$, where the sub-scripts indicate the word position in their corresponding sentences, then a movement

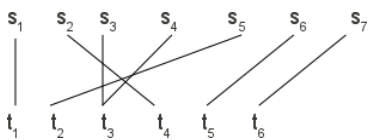

Figure 1: Word alignment between two sentences

of $d = 5 - 1 = 4$ words has taken place and its cost should be double than a movement of $d = 2$ words. A visual representation of these phrases can be seen in Figure 1

Besides the traditional distance-based reordering mentioned before, state-of-the-art systems implement an additional lexicalized reordering model (Tillman, 2004). The lexicalized reordering model classifies phrases by the movement they make relative to the previous used phrase, i.e., for each phrase the model learns how likely it is followed by the previous phrase (monotone), swapped with it (swap) or not connected at all (discontinuous). For instance, considering again sub-scripts as word positions in their corresponding sentences, in Figure 1 the bilingual phrases $(s_1 \rightarrow t_1)$ and $(s_5 \rightarrow t_2)$ are not connected, $(s_7 \rightarrow t_6)$ is followed by $(s_6 \rightarrow t_5)$ and $(s_2 \rightarrow t_4)$ is swapped with $(s_3\ s_4 \rightarrow t_3)$.

## 5.1 Direct System

Our direct system uses the phrase-based translation approach (Koehn et al., 2003). The basic idea is to segment the given source sentence $s$ into segments of one or more words, then each source segment is translated using a bilingual phrase obtained from the training corpus and finally compose the target sentence from these phrase translations. A bilingual phrase is a pair of $m$ source words and $n$ target words extracted from a parallel sentence that belongs to a bilingual corpus previously aligned by words. For extraction, we consider the words that are consecutive in both source and target sides and which are consistent with the word alignment. We consider a phrase is consistent with the word alignment if no word inside the phrase is aligned with one word outside the phrase.

Regarding the segmentation of the sentence in $K$ phrases, we assume that all possible segmentations (which are considered as a hidden variable $M$) have the same probability $\alpha(t)$:

$$P\left(s|t\right) \quad = \quad \sum_M P(s, M|t) \qquad (4)$$

$$= \quad \sum_M P(M|t)P(\bar{s}|\bar{t}) \qquad (5)$$

$$= \quad \alpha(t) \sum_M P(\bar{s}|\bar{t}) \qquad (6)$$

Then, we consider only monotone translations so the phrase $\bar{s_k}$ is produced by $\bar{t_k}$ (Zens, Och, & Ney, 2002).

$$P\left(\bar{s}|\bar{t}\right) \quad = \quad \prod_{k=1}^{K} p(\bar{s_k}, \bar{t_k}) \qquad (7)$$

Finally, phrase translation probabilities are estimated as relative frequencies over all bilingual phrases in the corpus.

$$p\left(s|t\right) \;=\; \frac{N\left(s,t\right)}{N\left(t\right)} \tag{8}$$

where $N\left(s,t\right)$ counts the number of times the phrase $s$ is translated as $t$ and $N\left(t\right)$ the number of times the phrase in the target language appears in the training corpus.

## 5.2 Pivot-Based Systems

The cascaded approach handles the source–pivot and the pivot–target system independently. They are both built and tuned to improve their local translation quality and then composed to translate from the source language to the target language in two steps: first, the translation output from source to pivot is computed and then it is used to obtain the target translation output.

The pseudo-corpus approach translates the pivot section of the source–pivot parallel corpus to the target language using a pivot–target system built previously. Then, a source–target SMT system is built using the source side and the translated pivot side of the source–pivot corpus. The pseudo-corpus system is tuned using an original source–target development corpus, since we have it available.

The triangulation approach combines the source–pivot ($P(s|p)$ and $P(p|s)$) and pivot–target ($P(p|t)$ and $P(t|p)$) relative frequencies following the strategy proposed by Cohn & Lapata (2007) in order to build a source–target translation model. The translation probabilities are computed assuming the independence between the source and target phrases when given the pivot phrase.

$$P\left(s|t\right) \;=\; \sum_p P(s|p)P(p|t) \tag{9}$$

$$P\left(t|s\right) \;=\; \sum_p P(t|p)P(p|s) \tag{10}$$

where $s$, $t$, and $p$ represent phrases in the source, target and pivot language respectively.

The lexical weights are computed in a similar manner, following the strategy proposed by Cohn & Lapata (2007). This approach does not handle the lexicalized reordering and the other pivot strategies and therefore represents a limitation in its potential. Instead, a simple distance-based reordering is applied during decoding. This model gives a cost linear to the reordering distance. For instance, skipping over two words costs twice as much as skipping over one word.

Once the corresponding translation model have been obtained, the source–target system is tuned using the same original source–target development corpus mentioned in the previous approach.

## 6. Evaluation Framework

The following section introduces the details of the evaluation framework. We report the statistics of the Un corpus, a description of how we built the systems and the evaluation details.

### 6.1 Corpus Statistics

As far as we know, and as discussed in section 4, three parallel corpora are available for the Chinese–Spanish language pair: Btec, *Holy Bible* and Un.[4] The former was used in the 2008 Iwslt and complete experiments of pivot strategies are reported in works such as Bertoldi et al. (2008). The *Holy Bible* was used for the similar purposes by Henríquez, Banchs & Mariño (2010).

In this study we decide to use the Un corpus taking advantage of the fact that it is the largest corpus (among those three) and it contains the same sentences in six languages, therefore we can experiment with different pivot languages.

When experimenting with different pivot languages, in order to make the systems as comparable as possible, we first did a sentence selection over the corpus so all systems were built exactly with the same training, tuning and testing sets. This selection process was as follows:

1. All corpora were tokenized, using the standard tokenizer available in Moses (Koehn, Hoang, Birch, Callison-Burch, Federico, Bertoldi, Cowan, Shen, Moran, Zens, Dyer, Bojar, Constantin, & Herbst, 2007) for Spanish, English and French; ictclass (Zhang, Yu, Xiong, & Liu, 2003) for Chinese; and Mada+Tokan (Habash & Rambow, 2005) for Arabic.

2. The Spanish, English and French corpora were lowercased.

3. If a sentence had more than 100 words in any language, it was deleted from all corpora.

4. If a sentence pair had a word ratio larger than three for any Chinese–pivot or pivot–Spanish parallel corpora, it was deleted from all corpora.

5. To extract the tuning and test sets we identified all sentences occurring once in the corpora for all languages. The tuning and testing sets were drawn over these sentences to assure they do not appear in the training corpus. Additionally, from these sentences, we want to select those which differ most from the sentences in the training set and which have the lowest out-of-vocabulary rate. In order to do this, the perplexity over the English language model was computed on a sentence-by-sentence basis by using a leave-one-out strategy; then, we selected the two thousand sentences which had the highest perplexity and the lowest ratio of out-of-vocabulary words for constructing the tuning and testing sets. The highest perplexity criterion was used in order to avoid that tuning and test sentences were similar from the ones in the training set. The lowest out-of-vocabulary words criterion was used to minimize the number of out-of-vocabulary words in the tuning and test translation. The two criteria were used

---

4. During the review process of this paper, we have been aware of a new corpus KDE (K Desktop Environment), which is available from the recent OPUS project [5] sequentially, first we selected the sentences with the highest perplexity, and among them, we selected those with the lowest ratio of out-of-vocabulary words.

Table 2 shows the main statistics for all corpora used once divided for experimentation.

| Dataset | Lang. | Sent. | Words | Vocab. |
|---------|-------|-------|-------|--------|
| Train | Chinese | 58 | 1,700 | 17 |
| | Spanish | 58 | 2,300 | 20 |
| | English | 58 | 2,000 | 14 |
| | Arabic | 58 | 2,600 | 17 |
| | French | 58 | 2,300 | 18 |
| Dev. | Chinese | 1 | 33.0 | 3 |
| | Spanish | 1 | 43.4 | 5 |
| | English | 1 | 37.4 | 4.2 |
| | Arabic | 1 | 48.8 | 4.6 |
| | French | 1 | 44.1 | 5 |
| Test | Chinese | 1 | 33.7 | 3.8 |
| | Spanish | 1 | 44.2 | 5 |
| | English | 1 | 38.1 | 4.2 |
| | Arabic | 1 | 49.3 | 4.6 |
| | French | 1 | 44.9 | 5 |

Table 2: Un Corpus Statistics used for this research (all figures are given in thousands)

## 6.2 System Implementation and Evaluation Details

Our systems were build using revision 4075 of Moses (Koehn et al., 2007). For all systems, we used the default Moses parameters which includes the grow-diagonal-final-and word alignment symmetrization, the lexicalized reordering (where possible), relative frequencies, lexical weights and phrase bonus for the translation model (with phrases up to length 10), a 5-gram language model using Kneser-Ney smoothing and a word penalty model. Therefore, 14 different features are combined in equation (3). The language model was built using Srilm (Stolcke, 2002) version 1.5.12. The optimization was done using Mert (Och, 2003). For word aligning we used Giza++ (Och & Ney, 2000) version 1.0.5.

In order to evaluate the translation quality, we used Bleu (Papineni, Roukos, Ward, & Zhu, 2001), Ter (Snover, Dorr, Schwartz, Micciulla, & Makhoul, 2006) and Meteor (Banerjee & Lavie, 2005) automatic evaluation metrics.

Additionally, significance tests were performed to study when a system was better than the other. These tests followed the "pair bootstrap resampling" method presented by Koehn (2004): Given two translation outputs coming from two different systems, we created two new virtual test sets by drawing sentences with replacement from the translation outputs. Once we obtained them, we computed their Bleus and observed which system performs better. This procedure was repeated 1,000 times. At the end, if one of the systems out-performed the other 99% of the time, we concluded that it was indeed a better Bleu score with 99% statistical significance.

## 7. Chinese–Spanish Machine Translation Strategies

Given the different languages available in the Un corpora, we tested three different language pivots. Additionally, we compared the cascaded, pseudo-corpus and triangulation pivot strategies. Finally, we tried to combine the system outputs to improve the translation.

### 7.1 Experimenting with Different Pivot Languages

We built and compared several translation approaches in order to study the impact of the different pivot languages when translating from Chinese into Spanish. Moreover, we evaluated how the quality of pivot approaches differs from a direct system. We built the pivot systems using five of the languages available in the Un parallel corpus: English, Spanish, Chinese, Arabic and French, and we built the direct system on a Chinese–Spanish parallel corpus.

In particular, we experimented with the following Chinese-to-Spanish systems: the direct Chinese-to-Spanish system as a quality upper bound; three cascaded, three pseudo-corpus and three triangulation approaches, using English, Arabic and French as pivots. In order to build the pivot systems, we need the corresponding Chinese–pivot and pivot–Spanish systems.

Table 3 shows the Bleu, Ter and Meteor scores achieved with the intermediate systems trained with the Un Corpus that were later used to built the different pivot approaches. Meteor score for the Chinese-to-Arabic system is not shown as we did not have the postprocessing tools required for the language.

|                 | Bleu  | Ter   | Meteor |
|-----------------|-------|-------|--------|
| Chinese–English | 35.67 | 51.07 | 36.77  |
| Chinese–Arabic  | 46.11 | 56.12 | –      |
| Chinese–French  | 28.31 | 63.21 | 47.35  |
| English–Spanish | 51.22 | 32.02 | 70.12  |
| Arabic–Spanish  | 41.79 | 44.37 | 60.22  |
| French–Spanish  | 46.42 | 40.25 | 64.76  |

Table 3: Pivot Systems.

Table 4 shows the results for our Chinese-to-Spanish configurations with the Un corpus. We can see there that the best pivot system used the pseudo-corpus approach with English as the pivot language.

In Chinese-to-Spanish, the fact that the pseudo-corpus through English outperforms cascaded through English according to the Bleu score is not statistically significant, with a 99% confidence (Koehn, 2004). These results, however, are coherent with previous works using the same language pair (Bertoldi et al., 2008; Henríquez Q. et al., 2010) that also reported the pseudo-corpus strategy was better than the cascaded strategy. The cascaded and pseudo-corpus approaches through English are statistically significantly better than the triangulation approach, with a 99% confidence. To the best of our knowledge, reasons why one pivot approach is better than the other are not reported in the literature. Moreover, given that difference among approaches such as pseudo-corpus as cascaded approaches is not significant, it is better to perform experiments for each particular task and language pair.

| Languages | System | Bleu | Ter | Meteor | Pivot vocab. |
|---|---|---|---|---|---|
| Chinese–Spanish | direct | 33.06 | 57.32 | 53.96 | - |
| Chinese–English–Spanish | cascaded | 32.90 | 56.67 | 54.06 | 14k |
| Chinese–French–Spanish | cascaded | 30.37 | 60.33 | 50.96 | 18k |
| Chinese–Arabic–Spanish | cascaded | 28.88 | 60.37 | 50.15 | 17k |
| Chinese–English–Spanish | pseudo | 32.97 | 57.39 | 53.99 | 14k |
| Chinese–French–Spanish | pseudo | 32.61 | 57.43 | 53.55 | 18k |
| Chinese–Arabic–Spanish | pseudo | 32.23 | 57.47 | 53.27 | 17k |
| Chinese–English–Spanish | triangulation | 32.05 | 57.91 | 53.37 | 14k |
| Chinese–French–Spanish | triangulation | 30.41 | 59.70 | 51.51 | 18k |
| Chinese–Arabic–Spanish | triangulation | 30.61 | 59.53 | 51.43 | 17k |

Table 4: Chinese-to-Spanish cascaded, pseudo-corpus and triangulation approaches.

In all three approaches, according to the scores in table 4 English is the best pivot language, with a statistical signicance of 99%, which is coherent with the pivot–Spanish results in table 3.

As follows, we use a procedure to predict the most suitable pivot language and justify why a language may be a better pivot than another. For example, the pivot vocabulary sizes play an important role. Birch et al. (2008) concluded in their study that the target vocabulary size has a negative impact in the translation quality as measured with the Bleu score and it can be seen that Arabic and French have both a larger vocabulary size than English.

Apart from the vocabulary size, the research mentioned above also measured the success of machine translation in terms of word reordering, i.e., differences in word order that occur in a parallel corpus, which are mainly driven by syntactic differences between the languages.

In order to measure reordering in translation they assumed that reordering only occurs between two adjacent blocks in the source side. This simplification allowed them to detect a extract all reordering in a deterministic way.

A *block* $A_s$ is defined by Birch et. al. (2008) as a segment of consecutive source words (source span) which is aligned to a set of target words. The target words also form a *block* $A_t$. With the definition of *block* set, they formally defined a reordering $r$ as two blocks $A$ and $B$ that are adjacent in the source, the relative order of the blocks in the source is reversed in the target and the reordering is consistent. A reordering between *blocks* $A_s$ and $B_s$ is consistent if the *block* $C_s$, consisting of the union of *blocks* $A_s$ and $B_s$, is also consistent. A *block* $A_s$ is said to be consistent if the span defined by its corresponding target *block* $A_t$ does not contain words that are aligned to source words outside of $A_s$. This definition of a consistent *block* is equivalent to the definition of a phrase in the phrase-based machine translation paradigm. Finally, the set of all reorderings $r$ in a sentence is defined as $R$ and it is unique for a given pair of sentences. Summarizing, this concept of reordering is equivalent to the swap movement described in the lexicalized reordering at the end of section 5.

| Languages | Source–Pivot | Pivot–Target | Average |
|---|---|---|---|
| Chinese–English–Spanish | 0.3955 | 0.2124 | 0.3039 |
| Chinese–French–Spanish | 0.6200 | 0.0170 | 0.3185 |
| Chinese–Arabic–Spanish | 0.6921 | 0.0908 | 0.3914 |

Table 5: Chinese-to-Spanish RQuantity metrics depending on the pivot used.

With those concepts defined, they developed a metric called RQuantity, defined as a sentence level metric which is then averaged over a corpus:

$$\text{RQuantity} = \frac{\sum_{r \in R} |r_{A_s}| + |r_{B_s}|}{I} \qquad (11)$$

where $R$ is the set of reorderings for a sentence, $I$ is the source sentence length, $A$ and $B$ are the two blocks involved in the reordering, $|r_{A_s}|$ is the size or span of block $A$ on the source side and $|r_{B_s}|$ is the size or span of block $B$ on the source side (Birch et al., 2008).

The objective of the RQuantity is to measure the amount of reordering we need when translating from a source language to a target language. The minimum RQuantity for a given sentence is 0 if the translation does not involve any word movement and its maximum is $(\sum_{i=2}^{I} i)/I$ when the words in the translation are inverted compared with their order in the source sentence.

We have computed the RQuantity for the different language pairs involved in our pivot approaches. It can be seen in table 5 that English appears as the best pivot because it has the lowest average RQuantity between the three, i.e. it is the pivot that needs the least amount of reordering in average to achieve the final translation. French and Arabic required less movements to translate into Spanish than English, but a lot of reordering is needed to obtain the first step from Chinese, hence penalizing the average. This result is coherent with the conclusion obtained by Birch et al. (2008), which says that the amount of reordering has also a negative impact in Bleu score.

These results support the intuitive idea that English works as a good intermediate step between Chinese and Spanish. Both French and Arabic have a vocabulary which is closer in size to Spanish and their reorderings are also more complex than English during the first step, making the source-to-pivot translation harder with these candidates. The gradual increase in difficulty (measured as target vocabulary size and reordering) presented in English seems to benefit the global result.

Nevertheless, it is also possible that most Un texts were authored in English, and then, translated into the other languages. This would also favour English as the best pivot language.

In order to observe the benefits of the pivot language against the direct translation, table 6 presents three examples where the Bleu scores of the pivot approach were better than those of the direct approach. Notice how some phrases that disappeared from the direct translation correctly appear on the pseudo-corpus approach.

| | |
|---|---|
| DIRECT | cuestiones como a que consideren seriamente la posibilidad de ratificar la tortura y otros tratos o penas crueles , inhumanos o degradantes |
| PSEUDO | como cuestiones a que consideren seriamente la posibilidad de ratificar **la convención contra** la tortura y otros tratos o penas crueles , inhumanos o degradantes |
| REF | considere seriamente la posibilidad de ratificar , con carácter prioritario , **la convención contra** la tortura y otros tratos o penas crueles , inhumanos o degradantes |
| EN REF | to seriously consider ratifying , as a matter of priority , **the convention against** torture and other cruel , inhuman or degrading treatment or punishment |
| DIRECT | habiendo examinado el segundo informe de la **comisión** y la recomendación que figura en él |
| PSEUDO | habiendo examinado el segundo informe de la **comisión de verificación de poderes** y las recomendaciones que figuran en él |
| REF | habiendo examinado el segundo informe de la **comisión de verificación de poderes** y la recomendación que figura en él |
| EN REF | having considered the second report of the **credentials committee** and the recommendation contained therein |
| DIRECT | pide al secretario general que prepare un informe sobre la aplicación de esta resolución **a la asamblea general** , quincuagésimo sexto período de sesiones |
| PSEUDO | pide al secretario general que prepare un informe sobre la aplicación de la presente resolución **para su examen por la asamblea general** en su quincuagésimo sexto período de sesiones |
| REF | pide al secretario general que prepare un informe sobre la aplicación de la presente resolución , **que será examinado por la asamblea general** en su quincuagésimo sexto período de sesiones |
| EN REF | requests the secretary-general to prepare a report on the implementation of the present resolution **for consideration by the general assembly** at its fifty-sixth session . |

Table 6: Chinese-to-Spanish examples for which the pseudo-corpus system (through English) is better than the direct system. EN REF is the English reference of the sentence

## 7.2 Pivot Combination

Using the 1-best translation output from the different pivot strategies, we built an $n$-best list and computed the final translation using minimum Bayes risk (MBR) (Kumar & Byrne, 2004).

When translating a sentence $s$, we obtain a translation $t'$ which can then be evaluated against reference $t$ to measure the system's performance. MBR focuses on finding the best performance over all possible translations. To do so, it uses a loss function LF$(t, t')$ that measures the loss of obtaining hypothesis $t'$ instead of the real translation $t$. The Bayes Risk is defined as the expected value of the loss function over all possible hypotheses $t'$ and translations $t$.

$$\text{E(LF)} = \sum_{t,t'} \text{LF}(t, t') p(t'|s) \qquad (12)$$

where $p(t'|s)$ is the translation probability of hypothesis $t'$ given the source sentence $s$ as obtained by the decoder, as an approximation of its real probability distribution.

The objective of finding the best performance over all possible translation is therefore to minimize the Bayes Risk. Given a loss function and a distribution, the decision rule that minimizes the Bayes Risk (Bickel & Doksum, 1977; Goel & Byrne, 2000) is given by:

$$\hat{t} = \arg\min_{t'} \sum_{t} \text{LF}(t, t') p(t'|s) \qquad (13)$$

Mbr has been used in literature both during decoding (Ehling, Zens, & Ney, 2007) and as a postprocess over a $n$-best list. For instance, this last approach was used by Khalilov et al. (2008) together with their cascaded approach in order to obtain the best Chinese–Spanish translation. The current version of the Moses toolkit includes both implementations.

The Mbr algorithm implemented in Moses as postprocess uses $1 - \text{Bleu}(t, t')$ as the loss function. In our experiment, we consider all our hypothesis as equally likely and therefore $p(t'|s)$ was a positive constant and therefore could be discarded. At the end, Mbr chooses the hypotheses $\hat{t}$ that fullfils:

$$\hat{t} = \arg\min_{t'} \left( \sum_{t \neq t'} 1 - \text{Bleu}(t, t') \right) \tag{14}$$

Different $n$-best lists were built to compare different Mbr outputs: a cascaded Mbr using all three pivot languages (hence $n = 3$, one hypothesis per pivot), a pseudo-corpus Mbr again using all three pivot languages ($n = 3$), a triangulation Mbr using all three languages ($n = 3$), a combination of cascaded, pseudo-corpus and triangulation outputs using two languages ($n = 6$, one hypothesis per pivot and strategy) and another using all of them ($n = 9$). It is important to mention that all $n$-best lists must have at least 3 hypothesis per sentence. Having only two hypothesis would not work as expected because the Loss Function would always choose the longest one, which can be explained by the definition of Bleu:

$$\text{Bleu}(t, t') = \exp \left( \sum_{n=1}^{N} \log \frac{p_n(t, t')}{N} \right) \gamma(t, t') \tag{15}$$

where $p_n(t, t')$ is the precision of $n$-grams in the hypothesis $t'$ with reference $t$; and $\gamma(t, t')$ is a brevity penalty disfavouring translation $t'$ if it is shorter than the reference $t$. Then $p_n(t, t') = p_n(t', t)$ and

$$(\forall t, t' : \text{length}(t) > \text{length}(t')) \tag{16}$$

$$1 - \text{Bleu}(t, t') \geq 1 - \text{Bleu}(t', t) \tag{17}$$

Tables 7 and 8 show the results of the different Chinese–Spanish output systems (from table 4) combined with the Mbr technique. From these tables, it can be observed that combinations from all pivot strategies obtained better results in all metrics than the direct approach. Only in the case of $Ar + Fr$, the combination was not statistically significantly better than the direct system in terms of Bleu score (with a 99% confidence).

The Mbr cascaded and triangulation approach (1st row, 2nd and 4th column, respectively, in table 8) did not outperform the direct system.

Finally, both $A\ All$ and $D+A\ All$ (which combine all languages and pivot system outputs from table 4 including or not the direct approach) are the best Chinese-to-Spanish systems.

We have not experimented on the reverse translation direction (from Spanish into Chinese) as we would be unable to assess subjective evaluations on the resulting translation outputs. However, in the reversed direction, our intuition is that the reordering difficulties will be then moved to the pivot–target step of the cascade system.

|       | All                  |
|-------|----------------------|
| En+Fr | 33.58*/56.34/54.48   |
| En+Ar | 33.53*/56.15/54.63   |
| Ar+Fr | 33.14/56.83/53.95    |

Table 7: Chinese-to-Spanish percent Bleu/Ter/Meteor scores for system combinations of two languages and all pivot approaches using Mbr. (*) statistically significant better Bleu than the direct system.

|       | Cascaded            | Pseudo              | Triangulation        | All                  |
|-------|---------------------|---------------------|----------------------|----------------------|
| A     | 32.66/57.27/53.34   | 33.30*/57.04/53.91  | 31.84/58.12/53.05    | 33.97*/56.00/54.87   |
| D+A   | 33.60*/56.72/54.38  | 33.77*/56.52/54.47  | 32.90/57.01/54.03    | 34.09*/55.88/55.02   |

Table 8: Chinese-to-Spanish percent Bleu/Ter/Meteor scores for system combinations of En + Fr + Ar languages (A), direct system (D) and pivot approaches using Mbr. (*) statistically significant better Bleu than the direct system.

Regarding the fact of English being the best pivot language for the task under consideration, we can argue that English might constitute a better intermediate step between Spanish and Chinese, rather than French or Arabic, based on the assumption of Spanish being closer to French (both are romance languages derived from Latin) and Arabic (the Iberian peninsula was occupied by Arabic culture for more than 500 years, so Spanish has strong influence from Arabic) than Chinese to French and Arabic. In this sense, English seems to represent an optimal intermediate point between Chinese and Spanish, in which the translation complexity is divided in two phases. Most of the reordering burden is resolved in the Chinese-to-English phase and most of the morphology generation burden is resolved in the English-to-Spanish phase. Thinking on translation-space as a non-conservative field, we can say that English is just in the middle of the way between Chinese and Spanish, while passing through French or Arabic implies a larger path by some kind of detour in the proximities of Spanish. This is just a conjecture, of course, but it nicely explains what we are observing. Definitively, more research is needed to better understand what is happening.

## 8. Conclusions

This work provided a brief survey in the state-of-the-art of Chinese–Spanish Smt. First of all, this language pair is of great interest both economically and culturally if we take into account the high number of Chinese and Spanish speakers. Besides, statistical machine translation is the most popular approach in the field of Mt given that has shown great quality in all the international evaluation campaigns such as Nist (2009) and Wmt (2012). The main points covered in our study were:

- There are mainly three Chinese–Spanish parallel corpora (Btec, *Holy Bible* and Un) that are freely available for research purposes.

- English is the best pivot language for conducting Chinese-to-Spanish translations compared to languages such as French or Arabic. The system built using English as pivot was significantly better than the ones built with French or Arabic, with a 99% confidence in both comparisons.

- The preference for a pivot language appears to be correlated with other proposed translation-quality prediction metrics such as the differences in vocabulary sizes and the amount of reordering. According to the above conclusion, the best pivot language is English because it has the lowest increase in vocabulary size and the lowest increase in reordering complexity.

- No significant difference is found among the best cascaded and pseudo-corpus pivot approaches, but the pseudo-corpus strategy is the best pivot strategy for Chinese-to-Spanish. Additionally, pseudo-corpus and cascaded approaches are significantly better than the triangulation approach.

- The output combination using MBR is able to improve the direct system in 1 BLEU point in the best case. This improvement is significantly better with a 99% confidence and is coherent with improvements in all other evaluation metrics studied.

As future research we plan to work on the problem of automatically extracting parallel corpus from comparable corpora collected from the web. Additionally, we intend to develop a freely available Chinese–Spanish translation system which would allow for collecting user feedback. Then, we will work on special techniques to incorporate this knowledge in the SMT system.

## Acknowledgments

The authors would like to specially thank the reviewers for their comments that helped a lot to improve this work. Additionally, the authors would like to thank the Universitat Politècnica de Catalunya and the Institute for Infocomm Research for their support and permission to publish this research.

This work has been partially funded by the Seventh Framework Program of the European Commission through the International Outgoing Fellowship Marie Curie Action (IMTraP-2011-29951); and by the Spanish Ministry of Economy and Competitiveness through the FPI Scholarship BES-2008-003851 for Ph.D. students under the AVIVAVOZ project (TEC2006-13694-C03-01); and the BUCEADOR project (TEC2009-14094-C04-01).